\documentclass{opt2020} 
\usepackage{booktabs} 
\usepackage{amsmath}
\usepackage{amssymb}
\usepackage{bbm}
\usepackage{color}
\usepackage{float}
\usepackage{amsmath,stmaryrd,pict2e,picture}
\usepackage{tcolorbox}
\usepackage{nicefrac}
\usepackage[colorinlistoftodos,bordercolor=orange,backgroundcolor=orange!20,linecolor=orange,textsize=scriptsize]{todonotes}

\newcommand{\R}{\mathbb{R}^d}

\newcommand{\cS}{\mathcal{S}}
\newcommand{\cD}{\mathcal{D}}
\newcommand{\sg}{{\sigma}}
\newcommand{\tk}{{\tau^k}}
\newcommand{\lk}{{\mathcal L^k}}

\newcommand{\norm}[1]{\left\| #1 \right\|}
\newcommand{\Exp}[1]{{\mathbb E}\left[#1\right]}

\newtheorem{assumption}{Assumption}

\title[Adaptive Learning of the Optimal Batch Size of SGD]{Adaptive Learning of the Optimal Batch Size of SGD}

\optauthor{\Name{Motasem Alfarra}\Email{motasem.alfarra@kaust.edu.sa} \\\Name{Slavom\'ir Hanzely}\Email{slavomir.hanzely@kaust.edu.sa}\\\Name{Alyazeed Albasyoni}\Email{alyazeed.albasyoni@kaust.edu.sa} \\\Name{Bernard Ghanem}\Email{bernard.ghanem@kaust.edu.sa} \\\Name{Peter Richt\'arik}\Email{peter.richtarik@kaust.edu.sa}\\
  }

\begin{document}

\maketitle

\vspace{-1.2cm}
\begin{abstract}
Recent advances in the theoretical understanding of SGD \cite{SGD_general_analysis} led to a formula for the optimal   batch size  minimizing the number of effective data passes, i.e., the number of iterations times the   batch size. However, this formula is of no practical value as it depends on the knowledge of the variance of the stochastic gradients evaluated at the optimum.  In this paper we design a practical SGD method capable of learning the optimal   batch size adaptively throughout its iterations for strongly convex and smooth functions. Our method does this provably, and in our experiments with synthetic and real data robustly exhibits nearly optimal behaviour; that is, it works as if the optimal  batch size was known a-priori. Further,  we generalize our method to several new  batch strategies not considered in the literature before, including a sampling suitable for distributed implementations.
\end{abstract}

\section{Introduction}

Stochastic Gradient Descent (SGD), in one disguise or another, is undoubtedly the backbone of modern systems for training supervised machine learning models~\cite{robbins1951stochastic, nemirovski2009robust, bottou2010large}. The method earns its popularity due to its superior performance on very large datasets where more traditional methods such as gradient descent (GD), relying on a pass through the entire training dataset before adjusting the model parameters, are simply too slow to be useful. In contrast, SGD in each iteration uses a small portion of the training data only (a  batch) to adjust the model parameters, and this process repeats until a model of suitable quality is found. 
In practice,  batch SGD is virtually always  applied to a finite-sum problem of the form $\textstyle  x^* = \arg \min \limits_{x\in \R} \frac{1}{n} \sum \limits_{i = 1}^n f_i(x),$
where $n$ is the number of training data and $f(x)=\frac{1}{n} \sum_{i = 1}^n f_i(x)$ represents the average loss, i.e.\ empirical risk, of model $x$ on the training dataset. With this formalism in place, a generic  batch SGD method performs the iteration $\textstyle x^{k+1} = x^k - \gamma^k \sum \limits_{i\in S^k} v_i^k \nabla f_i(x^k),$
where $S^k \subseteq \{1,2,\dots,n\}$ is the  batch considered in iteration $k$ and $v_1^k,\dots,v_n^k$ are appropriately chosen scalars. Often in practice, and almost invariably in theory, the  batch $S^k$ is chosen at random according to some fixed probability law, and the scalars $v_i^k$ are chosen to ensure that $g^k=\sum \limits_{i\in S^k} v_i^k \nabla f_i(x^k)$
is an unbiased estimator of the gradient $\nabla f(x^k)$. One standard choice is to fix a  batch size $\tau \in \{1,2,\dots,n\}$, and pick $S^k$ uniformly from all subsets of size $\tau$. Another option is to partition the training dataset into $n/\tau$ subsets of size $\tau$, and then in each iteration let $S^k$ to be one of these partitions, chosen with some probability, e.g., uniformly.

\textbf{Contributions}
{\bf Effective online learning of the optimal  batch size.} We make a step towards the development of a practical variant of optimal  batch SGD, aiming to learn the optimal  batch size $\tau^*$ on the fly. To the best of our knowledge, our method (Algorithm~\ref{alg:sgd_dyn_general}) is the first variant of SGD able to learn the optimal  batch.
\textbf {Sampling strategies.} We do not limit our selves to the uniform sampling strategy we used for illustration purposes above and develop closed-form expressions for the optimal  batch size for several other sampling techniques. Our adaptive method works well for all of them.
\textbf {Convergence theory.} We prove that our adaptive method converges, and moreover learns the optimal  batch size.
\textbf {Practical robustness.} We show the algorithm's robustness by conducting extensive experiments using different sampling techniques and different machine learning models on both real and synthetic datasets.

\textbf{Related work}
A stream of research attempts to boost the performance of SGD in practice is tuning its hyperparameters such as learning rate, and  batch size while training. In this context, a lot of work has been done in proposing various learning rate schedulers \cite{schumer1968adaptive, mathews1993stochastic,barzilai1988two,zeiler2012adadelta,tan2016barzilai,adaptive_step_size}. \citet{adaptive-mini-batch} showed that one can reduce the variance by increasing the  batch size without decreasing step-size (to maintain the constant signal to noise ratio). Besides, \citet{dont_decrease_step_size} demonstrated the effect of increasing the batch size instead of decreasing the learning rate in training a deep neural network. However, most of these strategies are based on empirical results only. \citet{you2017large,you2017scaling} show empirically the advantage of training on large  batch size, while \citet{masters2018revisiting} claim that it is preferable to train on smaller one.

\textbf{SGD Overview.} To study batch SGD for virtually all (stationary) subsampling rules, we adopt the stochastic reformulation paradigm for finite-sum problems proposed in \cite{SGD_general_analysis}. The random vector $v \in \mathbb{R}^n$ is sampled from a distribution $\cD$ and satisfies $\mathbb{E}_{\mathcal{D}}[v_i] = 1$.
Typically, the vector $v$ is defined by first choosing a random  batch $\cS^k\subseteq \{1,2,\dots,n\}$, then defining $v_i^k  = 0$ for $i\notin \cS^k$, and choosing $v_i^k$ to an appropriate value for $i\notin \cS^k$ in order to make sure the stochastic gradient $\nabla f_{v^k}(x^k)$ is unbiased. 
In this work, we consider two particular choices of the probability law governing the selection of $\cS^k$.



\textbf{$\tau-$partition nice sampling} In this sampling, we divide the training set into partitions $\mathcal C_j$ (of possibly different sizes $n_{\mathcal C_j}$), and each of them has at least a cardinality of $\tau$. At each iteration, one of the sets $\mathcal C_j$ is chosen with probability $q_{\mathcal C_j}$, and then $\tau-$nice sampling (without replacement) is applied on the chosen set. For each subset $C$ cardinality $\tau$ of partition $C_j$ cardinality $n_{C_j}$, $\mathbb P[v_C=\mathbbm 1_{C \in \mathcal S}]=q_j/\binom {n_{C_j}} \tau$. 
\textbf{$\tau-$partition independent sampling} Similar to $\tau-$partition nice sampling, we divide the training set into partitions $\mathcal C_j$, and each of them has at least a cardinality of $\tau$. At each iteration one of the sets $\mathcal C_j$ is chosen with probability $q_{\mathcal C_j}$, and then $\tau-$independent sampling is applied on the chosen set. For each element $i$ of partition $C_j$, we have $\mathbb P[v_i=\mathbbm 1_{C \in \mathcal S}]=q_jp_i$. 
The stochastic formulation naturally leads to the following concept of expected smoothness.
\vspace{-0.2cm}
\begin{assumption}
\label{as:expected_smoothness}
The function $f$ is $\mathcal{L}-$smooth with respect to a datasets $\mathcal{D}$ if there exist $\mathcal{L} > 0$ with
\begin{equation}\label{eq:exp_smoothn_cL}
\mathbb{E}_\mathcal{D}\left[\left\| \nabla f_v(x) - \nabla f_v(x^*) \right\|^2\right] \leq 2 \mathcal{L}(f(x) - f(x^*)). 
\end{equation}
\end{assumption}

\vspace{-0.75cm}
\begin{assumption}
\label{as:gradient_noise}
The gradient noise $\sg = \sg(x^*)$, where 
$    \sg(x) := \mathbb{E}_\mathcal{D}[\left\| \nabla f_v(x) \right\|^2  ],
$
is finite.
\end{assumption}
\vspace{-0.6cm}
\begin{theorem}
\label{th:convergence_general}
Assume $f$ is $\mu-$strongly convex and Assumptions \ref{as:expected_smoothness}, \ref{as:gradient_noise} are satisfied. For any $\epsilon > 0$, if the learning rate $\gamma$ is set to be
\vspace{-0.2cm}
\begin{equation} \label{step_size}
    \gamma = \tfrac 1 2 \min \left\{ \tfrac{1}{\mathcal{L}}, \tfrac{\epsilon \mu}{2\sg} \right\} \quad \text{and} \quad k \geq \tfrac 2 {\mu} \max\left\{ \mathcal{L}, \tfrac{2\sg}{\epsilon \mu} \right\} \log\left( \tfrac{2 \|x^0 - x^*\|^2}{\epsilon} \right) \,\, \text{then}\,\, \mathbb{E} \left\| x^k - x^* \right\|^2  \leq \epsilon
\end{equation}
\vspace{-0.2cm}
\end{theorem}


\vspace{-1.25cm}
\section{Deriving Optimal  Batch Size} \label{sec:analysis}
After giving this thorough introduction to the stochastic reformulation of SGD, we can move on to study the effect of the  batch size on the total iteration complexity. In fact, for each sampling technique, the  batch size will affect both the expected smoothness $\mathcal{L}$ and the gradient noise $\sg$. This effect reflects on the number of iterations required to reach to $\epsilon$ neighborhood around the optimum.

\textbf{Formulas for $\mathcal L$ and $\sigma$}
Before proceeding, we establish some terminologies. In addition of having $f$ to be $L-$smooth, we also assume each $f_i$ to be $L_i-$smooth.
In $\tau-$partition samplings (both nice and independent), let $n_{\mathcal C_j}$ be number of data-points in the partition $\mathcal C_j$, where $n_{\mathcal C_j} \geq \tau$. Let $L_{\mathcal C_j}$ be the smoothness constants of the function $f_{\mathcal C_j} = \tfrac{1}{n_{\mathcal C_j}}\sum_{i \in \mathcal C_j} f_i$. Also, let $\overline L_{\mathcal C_j} = \tfrac{1}{n_{\mathcal C_j}} \sum_{i\in\mathcal C_j} L_i$ be the average of the Lipschitz smoothness constants of the functions in partition $\mathcal C_j$. In addition, let $h_{\mathcal C_j}(x) = \norm{\nabla f_{\mathcal{C}_j}(x)}^2$ be the norm of the gradient of $f_{\mathcal{C}_j}$ at $x$. Finally, let $\overline {h}_{\mathcal C_j}(x) = \tfrac {1} {n_{\mathcal C_j}} \sum_{i \in C_j} h_i(x)$. For ease of notation, we will drop $x$ from all of the expression since it is understood from the context $(h_i = h_i(x))$. Also, superscripts with $(*,k)$ refer to evaluating the function at $x^*$ and $x^k$ respectively $(e.g.\,\, h_i^* = h_i(x^*))$.
Now we introduce our key lemma, which gives an estimate of the expected smoothness for different sampling techniques.

\begin{lemma} \label{le:L}
For the considered samplings, the expected smoothness constants $\mathcal L$ can be upper bounded by $\mathcal L(\tau)$ (i.e. $\mathcal L \leq \mathcal L(\tau)$), where $\mathcal L(\tau)$ is expressed as follows

(i) For $\tau$-partition nice sampling,$\mathcal L (\tau) = \tfrac{1}{n\tau} \max_{\mathcal C_j} \tfrac{n_{\mathcal C_j}}{q_{\mathcal C_j}(n_{\mathcal C_j}-1)} 
\Big[(\tau-1) L_{\mathcal C_j}n_{\mathcal C_j}
+ (n_{\mathcal C_j}-\tau) \max_{i \in{\mathcal C_j}}{L_i} \Big].$


(ii) For $\tau$-partition independent sampling, we have: $\mathcal L (\tau) = \tfrac{1}{n}\max_{\mathcal C_j}{\tfrac{n_{\mathcal C_j} L_{\mathcal C_j}}{q_{\mathcal C_j}} + \max_{i \in C_j}{\tfrac{L_i(1-p_{i})}{q_{C_j}p_{i}}}}.$


For the considered samplings, the gradient noise is given by $\sg (x^*, \tau)$, where

(i) For $\tau$-partition nice sampling,$\sg(x,\tau) = \tfrac{1}{n^2\tau}\sum_{\mathcal C_j} \tfrac{n_{\mathcal C_j}^2}{q_{\mathcal C_j}(n_{\mathcal C_j}-1)} \Big[(\tau-1)  h_{\mathcal C_j}n_{\mathcal C_j} 
 + (n_{\mathcal C_j}-\tau) \overline {h}_{\mathcal C_j}\Big].$


(ii) For $\tau-$partition independent sampling, we have $\mathcal \sg(x,\tau) = \tfrac{1}{n^2}\sum_{\mathcal C_j}\tfrac{n_{\mathcal C_j}^2  h_{\mathcal C_j} + \sum_{i \in C_j}{\tfrac{1-p_{i}}{p_{i}} h_i}}{q_{C_j}}$


\end{lemma}



\textbf{Optimal  Batch Size}
Our goal is to estimate total iteration complexity as a function of $\tau$. In each iteration, we work with $\tau$ gradients, thus we can lower bound on the total iteration complexity by multiplying lower bound on iteration complexity \eqref{step_size} by $\tau$.
We can apply similar analysis as in \cite{SGD_general_analysis}. 
Since we have estimates on both the expected smoothness constant and the gradient noise in terms of the  batch size $\tau$, we can lower bound total iteration complexity \eqref{step_size} as
$ T(\tau) = \tfrac 2 {\mu} \max \left\{ \tau\mathcal L(\tau), \tfrac{2}{\epsilon \mu}\tau \sg(x^*,\tau) \right\} \log\left( \tfrac{2 \|x^0 - x^*\|^2}{\epsilon} \right).$
Note that if we are interested in minimizer of $T(\tau)$, we can drop all constant terms in $\tau$. Therefore, optimal  batch size $\tau^*$ minimizes $\max \left\{ \tau\mathcal L(\tau), \tfrac{2}{\epsilon \mu}\tau \sg(x^*,\tau) \right\}$. It turns out that all $\tau \mathcal L(\tau)$, and $\tau \sg(x^*,\tau)$ from Lemma \ref{le:L}, are piece-wise linear functions in $\tau$, which is cruicial in helping us find the optimal $\tau^*$ that minimizes $T(\tau)$ which can be accomplished through the following theorem.

\begin{theorem}\label{le:tau_star}
For $\tau$-partition nice sampling and $\tau-$partition independent sampling with $p_i = \frac{\tau}{n_{\mathcal C_j}}$, the optimal  batch size is $\tau(x^*)$, where $\tau(x)$ is given by
\begin{equation*} \label{eq:tau_part_nice}
    \min_{\mathcal C_r}\tfrac{
    \tfrac{nn_{\mathcal C_r}^2}{e_{\mathcal{C}_r}} (L_{\mathcal C_r}-L_{\max}^{\mathcal C_r}) + \tfrac{2}{\epsilon \mu} \sum_{\mathcal C_j} \tfrac{n_{\mathcal C_j}^3}{e_{\mathcal C_j}}\left(\overline {h}_{\mathcal C_j}- h_{\mathcal C_j}\right)}
    {\tfrac {nn_{\mathcal{C}_r}} {e_{\mathcal C_r}}(n_{\mathcal C_r}L_{\mathcal C_r}-L_{\max}^{\mathcal{C}_r}) + \tfrac{2}{\epsilon \mu}\sum_{\mathcal C_j} 
    \tfrac{n_{\mathcal C_j}^2 }{e_{\mathcal C_j}}({\overline {h}_{\mathcal C_j} -n_{\mathcal C_j}  h_{\mathcal C_j} })}, \,\,\min_{\mathcal C_r}{\tfrac{
    \tfrac{2}{\epsilon \mu}
    \sum_{\mathcal C_j} \tfrac{n_{\mathcal C_j}^2} { q_{\mathcal C_j}} \overline {h}_{\mathcal C_j} - \tfrac{n}{q_{\mathcal C_r}} L_{\max}^{\mathcal{C}_r} }
    {
    \tfrac{2}{\epsilon \mu}\sum_{\mathcal C_j} \tfrac {n_{\mathcal C_j}} { q_{\mathcal C_j}}
    (\overline {h}_{\mathcal C_j} - n_{\mathcal C_j} h_{\mathcal C_j})
    +
    \tfrac n {q_{\mathcal C_r}} (n_{\mathcal C_r}L_{\mathcal C_r}-L_{\max}^{\mathcal C_r})}},
\end{equation*}
respectively, if  $\quad\sum_{\mathcal C_j} \tfrac{n_{\mathcal C_j}^2}{e_j} ( h_{\mathcal C_j}^*n_{\mathcal C_j}-\overline {h}^*_{\mathcal C_j}) \le 0\, $ for $\tau$-partition nice sampling, and $\quad\sum_{\mathcal C_j} \tfrac{n_{\mathcal C_j}}
    { q_{\mathcal C_j}} (n_{\mathcal C_j}  h_{\mathcal C_j}^* - \overline {h}^*_{\mathcal C_j}) \le 0\,$ for $\tau-$partition independent sampling, where  $e_{\mathcal C_k}=q_{\mathcal C_k} (n_{\mathcal C_k} -1)$ and  $L_{\max}^{\mathcal{C}_r}=\max_{i \in \mathcal C_r}{L_i}$. Otherwise: $\tau(x^*)=1$.

\end{theorem}


\section{Proposed Algorithm} \label{sec:algo}



The theoretical analysis gives us the optimal   batch size for each of the proposed sampling techniques. However, we are unable to use these formulas directly since all of the expressions of optimal   batch size depend on the knowledge of $x^*$ through the values of $h_i^* \forall i \in [n]$. Our algorithm overcomes this problem by estimating the values of $h_i^*$ at every iteration by $h_i^k$. Although this approach seems to be mathematically sound, it is costly because it requires passing through the whole training set every iteration. Alternatively, a more practical approach is to store $h_i^0=h_i(x^0)\,\, \forall i \in [n]$, then set $h_i^k = \norm{\nabla f_i(x^k)}^2\,\, for\,\, i\in\mathcal S_k$ and $h_i^k = h_i^{k-1} \,\,for\,\, i \notin \mathcal S_k$, where $\mathcal S_k$ is the set of indices considered in the $k^{th}$ iteration. In addition to storing an extra $n$ dimensional vector, this approach costs only computing the norms of the stochastic gradients that we already used in the SGD step. Both options lead to convergence in a similar number of epochs, so we let our proposed algorithm adopt the second (more practical) option of estimating $h_i^*$.

In our algorithm, for a given sampling technique, we use the current estimate of the model $x^k$ to estimate the sub-optimal   batch size $\tk := \tau(x^k)$ at the $k^{\text{th}}$ iteration. Based on this estimate, we use Lemma \ref{le:L} in calculating an estimate for both the expected smoothness  $\mathcal{L}(\tau^k)$ and the noise gradient $\sg(x^k, \tau^k)$ at that iteration. After that, we compute the step-size $\gamma^k$ and finally conduct a SGD step. The summary can be found in Algorithm \ref{alg:sgd_dyn_general}. For theoretical convergence purposes, we cap $\sg^k$ by a positive constant $C$, and we set the learning rate at each iteration to $\gamma^k \leftarrow \frac 1 2 \min \left\{\frac 1 {\lk}, \frac{\epsilon \mu}{\min\{C,2\sg^k\}} \right\}$. This way, learning rates generated by Algorithm \ref{alg:sgd_dyn_general} are bounded by positive constants $\gamma_{\max}= \tfrac 1 2\max_{\tau \in [n]} \left\{ \tfrac 1 {\mathcal L(\tau)} \right\}$ and $ \gamma_{\min}= \frac 1 2\min \left\{ \min_{\tau \in [n]} \left\{\tfrac 1 {\mathcal L(\tau)} \right\}, \tfrac{\epsilon \mu} C \right\} $.


\begin{algorithm2e}[t]
\caption{SGD with Adaptive  Batch size}
\label{alg:sgd_dyn_general}
\begin{algorithmic}
\STATE \textbf{Input:} Smoothness constants $L$, $L_i$, strong convexity constant $\mu$, target neighborhood $\epsilon$, Sampling Strategy $S$, initial point $x^0$, variance cap $C \geq 0$. \textbf{Initialize:} Set $k=0$
\WHILE{not converged}
    \STATE Set $\tau^k \leftarrow \tau(x^k)$,  $\quad \lk \leftarrow \mathcal L(\tau^k)$, $\quad \sg^k \leftarrow \sg(x^k, \tau^k)$, $\quad \gamma^k \leftarrow \frac 1 2 \min \left\{\frac 1 {\lk}, \frac{\epsilon \mu}{\min(C,2\sg^k)} \right\}$ 
    \STATE Sample $v_k$ from $S$ and Do SGD step: $x^{k+1} \leftarrow x^k - \gamma^k \nabla f_{v_k} (x^k)$
\ENDWHILE. $\quad$ \textbf{Output:} $x^k$
\end{algorithmic}
\end{algorithm2e}

\begin{figure*}[t]
\begin{center}
\centerline{
\includegraphics[width=0.25\columnwidth]{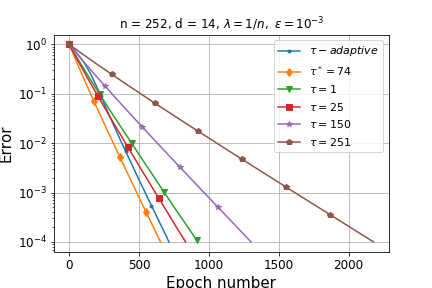}
\includegraphics[width=0.25\columnwidth]{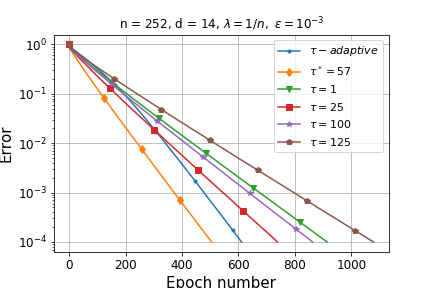}
\includegraphics[width=0.25\columnwidth]{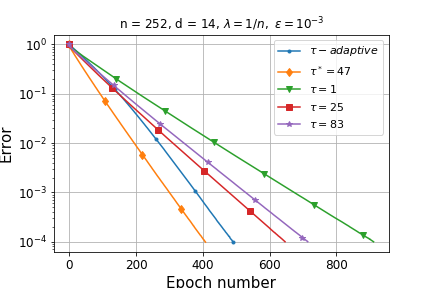}
\includegraphics[width=0.25\columnwidth]{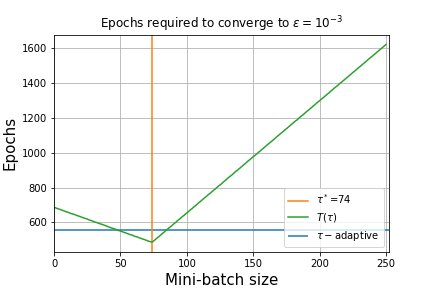}
}
\centerline{
\includegraphics[width=0.25\columnwidth]{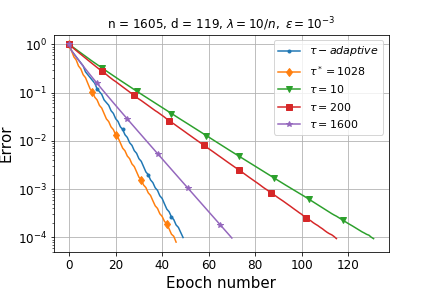}
\includegraphics[width=0.25\columnwidth]{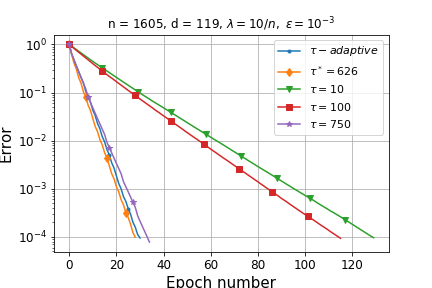}
\includegraphics[width=0.25\columnwidth]{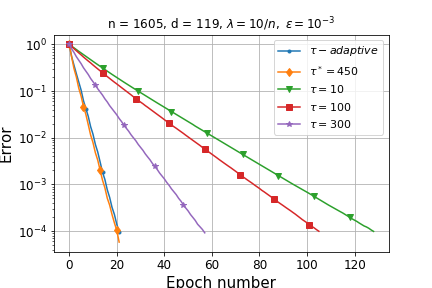}
\includegraphics[width=0.25\columnwidth, height = 72 pt]{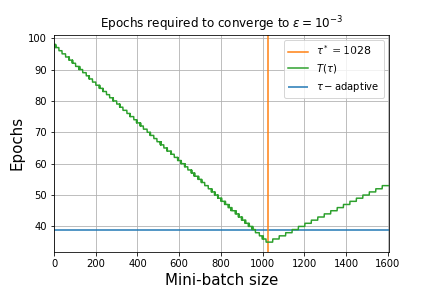}
}
\caption{\textbf{Convergence of ridge and logistic regression} using $\tau-$partition nice sampling on \emph{bodyfat} dataset (first row) and $\tau-$partition independent sampling on \emph{a1a} dataset (second row).
}
\label{fig:ridge_nice}
\end{center}
\vspace{-0.75cm}
\end{figure*}

\vspace{-0.2cm}
\begin{theorem}\label{th:convergence_our}
Assume $f$ is $\mu-$strongly convex and assumptions \ref{as:expected_smoothness}, and \ref{as:gradient_noise} hold. 
Then the iterates of Algorithm \ref{alg:sgd_dyn_general} satisfy: $\mathbb{E}\left\|x^k - x^*\right\|^2 \leq \left(1-\gamma_{\min}\mu\right)^k\left\|x^0-x^*\right\|^2 + R,$

\end{theorem}

where $R = \tfrac{2\gamma_{\max}^2\sg^*}{\gamma_{\min}\mu}$. 
Theorem \ref{th:convergence_our} guarantees the convergence of the proposed algorithm. Although there is no significant theoretical improvement here compared to previous SGD results in the fixed   batch and learning rate regimes, we measure the improvement to be significant in practice. 

\textbf{Convergence of $\tk$ to $\tau^*$}. The motivation behind the proposed algorithm is to learn the optimal   batch size in an online fashion so that we get to $\epsilon-$neighborhood of the optimal model with the minimum number of epochs. For simplicity, let's assume that $\sigma^*=0$. As $x^k \rightarrow x^*$, then $h_i^k=\nabla f_i(x^k) \rightarrow \nabla f_i(x^*) = h_i^*$, and thus  $\tk \rightarrow \tau^*$. 
In Theorem \ref{th:convergence_our}, we showed the convergence of $x^k$ to a neighborhood around $x^*$. Hence the theory predicts that our estimate of the optimal   batch $\tk$ will converge to a neighborhood of the optimal   batch size $\tau^*$.

\section{Experiments} \label{sec:experiments}
In this section, we compare our algorithm to fixed   batch size SGD in terms of the number of epochs needed to reach a pre-specified neighborhood $\epsilon/10$. In the following results, we capture the convergence rate by recording the relative error $(\nicefrac{\norm{x^k - x^*}^2}{\norm{x^0 - x^*}^2})$ where $x^0$ is drawn from a standard normal distribution $\mathcal{N}(0, \mathbf{I})$. We also report the number of training examples $n$, the dimension of the machine learning model $d$, regularization factor $\lambda$, and the target neighborhood $\epsilon$ above each figure.
We consider the problems of regularized ridge and logistic regression where each $f_i$ is strongly convex and L-smooth, and $x^*$ can be known a-priori.Specifically, we want to $\min_{x \in \mathbb R^d} f(x)$ where\vspace{-0.25cm}
\[\vspace{-0.25cm}
f_{\text{ridge}}(x) = \tfrac{1}{2n}\sum \limits_{i=1}^n \norm{a_i^T x - b_i}^2_2 + \tfrac{\lambda}{2}\norm{x}^2_2, \,\, f_{\text{logistic}}(x) = \tfrac{1}{2n}\sum \limits_{i=1}^n\log\left(1+\exp \left(b_i a_i^T x\right)\right) + \tfrac{\lambda}{2}\norm{x}^2_2
\]
where $(a_i, b_i) \sim \mathcal{D}$ are pairs of data examples from the training set.
For each of the considered problems, we performed experiments on real datasets from LIBSVM \cite{libsvm}. We tested our algorithm on ridge and logistic regression on \emph{bodyfat} and \emph{a1a} datasets in Figure \ref{fig:ridge_nice}. 
For these datasets, we considered $\tau-$partition independent and $\tau-$partition nice sampling with distributing the training set into one, two, and three partitions.  
Moreover, we take the previous experiments one step further by running a comparison of various fixed   batch size SGD, as well as our adaptive method with a single partition (last column of \ref{fig:ridge_nice}).
We plot the total iteration complexity for each   batch size, and highlight optimal   batch size obtained from our theoretical analysis, and how many epochs our adaptive algorithm needs to converge. This plot can be viewed as a summary of grid-search for optimal   batch size (throughout all possible fixed   batch sizes). 
Despite the fact that the optimal   batch size is nontrivial and varies significantly with the model, dataset, sampling strategy, and number of partitions, our algorithm demonstrated consistent performance overall. In some cases, it was even able to cut down the number of epochs needed to reach the desired error to a factor of six.

The produced figures of our grid-search perfectly capture the tightness of our theoretical analysis. In particular, the total iteration complexity decreases linearly up to a neighborhood of $\tau^*$ and then increases linearly. In addition, Theorem \ref{le:tau_star} always captures the empirical minimum of $T(\tau)$ up to a negligible error. 
Moreover, these figures show how close $T_\text{adaptive}$ is to the total iteration complexity using optimal   batch size $T(\tau^*)$.
Finally, in terms of running time,
our algorithm requires $0.2322$ ms per epoch, while running SGD with the optimal  batch size requires $0.2298$ ms.



\section{Acknowledgement}
This work was supported by the King Abdullah University of Science and Technology (KAUST) Office of Sponsored Research. The work of Motasem Alfarra and Bernard Ghanem was supported by Award No. OSR-CRG2019-4033.





\bibliography{references}

\newpage
\onecolumn
\appendix

\section{Proof of Lemma \ref{le:L}}
For the considered partition sampling, the indices $1, \dots, n$ are distributed into the sets $\mathcal C_1, \dots, \mathcal C_K$ with each having a minimum cardinality of $\tau$. We choose each set $\mathcal C_j$ with probability $q_{\mathcal{C}_j}$ where $\sum \limits_{j} q_{\mathcal C_j}=1$. Note that

\[
\mathbf P_{ij}=
\begin{cases}
0 & \text{if } i \in C_k, j \in C_l, k \neq l \\
q_k \frac{\tau(\tau-1)}{n_k(n_k-1)} & \text{if } i \neq j,  i, j \in C_k, |C_k|=\tau_k\\
q_k \frac {\tau} {n_k} & \text{if i=j}
\end{cases}.
\]

Therefore
\begin{eqnarray*}
\Exp{\norm{\nabla f_v(x) - \nabla f_v(y)}^2} &=& \frac{1}{n^2}\sum_{\mathcal C_k} \sum_{i,j \in \mathcal C_k} \frac{\mathbf P_{ij}}{p_ip_j} \Big \langle \nabla f_i(x)-\nabla f_i(y), \nabla f_j(x)-\nabla f_j(y) \Big \rangle \\
&=& \frac{1}{n^2} \sum_{\mathcal C_k} \sum_{i \neq j \in \mathcal C_k} \frac{\mathbf P_{ij}}{p_ip_j} \Big \langle \nabla f_i(x)-\nabla f_i(y),\nabla f_j(x)-\nabla f_j(y) \Big \rangle   \\
&\quad& + \frac{1}{n^2} \sum_{\mathcal C_k} \sum_{i \in \mathcal C_k} \frac{1}{p_i} \Big \langle \nabla f_i(x)-\nabla f_i(y),\nabla f_i(x)-\nabla f_i(y) \Big \rangle\\
&=& \frac{1}{n^2} \sum_{\mathcal C_k} \sum_{i \neq j \in \mathcal C_k} \frac{n_{\mathcal C_k}(\tau-1)}{q_{\mathcal C_k}\tau(n_{\mathcal C_k}-1)} \Big \langle \nabla f_i(x)-\nabla f_i(y),\nabla f_j(x)-\nabla f_j(y) \Big \rangle   \\
&\quad& + \frac{1}{n^2} \sum_{\mathcal C_k} \sum_{i \in \mathcal C_k} \frac{n_{\mathcal C_k}}{q_{\mathcal C_k}\tau} \Big \langle \nabla f_i(x)-\nabla f_i(y),\nabla f_i(x)-\nabla f_i(y) \Big \rangle\\
&=& \frac{1}{n^2} \sum_{\mathcal C_k} \frac{n_{\mathcal C_k}(\tau-1)}{q_{\mathcal C_k}\tau(n_{\mathcal C_k}-1)} \sum_{i \neq j \in \mathcal C_k} \Big \langle \nabla f_i(x)-\nabla f_i(y),\nabla f_j(x)-\nabla f_j(y) \Big \rangle   \\
&\quad& + \frac{1}{n^2}\sum_{\mathcal C_k} \frac{n_{\mathcal C_k}}{q_{\mathcal C_k}\tau} \sum_{i \in \mathcal C_k} \norm{\nabla f_i(x)-\nabla f_i(y)}^2\\
&=& \frac{1}{n^2} \sum_{\mathcal C_k} \frac{n_{\mathcal C_k}(\tau-1)}{q_{\mathcal C_k}\tau(n_{\mathcal C_k}-1)} \norm{\sum_{i \in \mathcal C_k} \nabla f_i(x)-\nabla f_i(y)}^2  \\
&\quad& + \frac{1}{n^2}\sum_{\mathcal C_k} \frac{n_{\mathcal C_k}(n_{\mathcal C_k}-\tau)}{q_{\mathcal C_k}\tau(n_{\mathcal C_k}-1)} \sum_{i \in \mathcal C_k} \norm{\nabla f_i(x)-\nabla f_i(y)}^2 \\
&\leq& \frac{1}{n^2} \sum_{\mathcal C_k} \frac{n_{\mathcal C_k}^3(\tau-1)}{q_{\mathcal C_k}\tau(n_{\mathcal C_k}-1)} 2L_{\mathcal C_k}D_{f_{\mathcal C_k}}(x, y)  \\
&\quad& + \frac{1}{n^2}\sum_{\mathcal C_k} \frac{n_{\mathcal C_k}(n_{\mathcal C_k}-\tau)}{q_{\mathcal C_k}\tau(n_{\mathcal C_k}-1)} \sum_{i \in \mathcal C_k} 2L_iD_{f_i}(x,y) \\
&\leq& \frac{1}{n^2} \sum_{\mathcal C_k} \frac{n_{\mathcal C_k}^3(\tau-1)}{q_{\mathcal C_k}\tau(n_{\mathcal C_k}-1)} 2L_{\mathcal C_k}D_{f_{\mathcal C_k}}(x, y)  \\
&\quad& + \frac{1}{n^2}\sum_{\mathcal C_k} \frac{n_{\mathcal C_k}^2(n_{\mathcal C_k}-\tau)}{q_{\mathcal C_k}\tau(n_{\mathcal C_k}-1)} 2\max_{i \in \mathcal C_k}{L_i}D_{f_{\mathcal C_k}}(x,y) \\
&=& \frac{1}{n^2} \sum_{\mathcal C_k} 2\frac{n_{\mathcal C_k}^2(\tau-1)L_{\mathcal C_k} + n_{\mathcal C_k}(n_{\mathcal C_k}-\tau)\max_{i \in \mathcal C_k}{L_i} }{q_{\mathcal C_k}\tau(n_{\mathcal C_k}-1)} n_{\mathcal C_k}D_{f_{\mathcal C_k}}(x, y)  \\
&\leq& 2\frac{1}{n}( \max_{\mathcal C_k}\frac{n_{\mathcal C_k}^2(\tau-1)L_{\mathcal C_k} + n_{\mathcal C_k}(n_{\mathcal C_k}-\tau)\max_{i \in \mathcal C_k}{L_i} }{q_{\mathcal C_k}\tau(n_{\mathcal C_k}-1)} n_{\mathcal C_k}) D_{f}(x,y),
\end{eqnarray*}
where $D_f(x,y) = f(x) - f(y) - \langle\nabla f(y), x-y\rangle$. 
Setting $y \leftarrow x^*$, leads to the desired upper bound of the expected smoothness which is given by
\begin{eqnarray*}
\mathcal L(\tau) = \frac{1}{n\tau}\left( \max_{\mathcal C_k}\frac{n_{\mathcal C_k}}{q_{\mathcal C_k}\left(n_{\mathcal C_k}-1\right)}\left(n_{\mathcal C_k}^2(\tau-1)L_{\mathcal C_k} + n_{\mathcal C_k}\left(n_{\mathcal C_k}-\tau\right)\max_{i \in \mathcal C_k}{L_i}\right)\right).
\end{eqnarray*}

Next, we derive a similar bound  for $\tau-$independent partition sampling.
\begin{eqnarray*}
\Exp{\norm{\nabla f_v(x) - \nabla f_v(y)}^2} &=& \frac{1}{n^2}\sum_{\mathcal C_k} \sum_{i,j \in \mathcal C_k} \frac{\mathbf P_{ij}}{p_ip_j} \Big \langle \nabla f_i(x)-\nabla f_i(y), \nabla f_j(x)-\nabla f_j(y) \Big \rangle \\
&=& \frac{1}{n^2} \sum_{\mathcal C_k} \sum_{i \neq j \in \mathcal C_k} \frac{\mathbf P_{ij}}{p_ip_j} \Big \langle \nabla f_i(x)-\nabla f_i(y),\nabla f_j(x)-\nabla f_j(y) \Big \rangle   \\
&\quad& + \frac{1}{n^2} \sum_{\mathcal C_k} \sum_{i \in \mathcal C_k} \frac{1}{p_i} \Big \langle \nabla f_i(x)-\nabla f_i(y),\nabla f_i(x)-\nabla f_i(y) \Big \rangle\\
&=& \frac{1}{n^2} \sum_{\mathcal C_k} \sum_{i \neq j \in \mathcal C_k} \frac{1}{q_{\mathcal C_k}} \Big \langle \nabla f_i(x)-\nabla f_i(y),\nabla f_j(x)-\nabla f_j(y) \Big \rangle   \\
&\quad& + \frac{1}{n^2} \sum_{\mathcal C_k} \sum_{i \in \mathcal C_k} \frac{1}{q_{\mathcal C_k}p_i} \Big \langle \nabla f_i(x)-\nabla f_i(y),\nabla f_i(x)-\nabla f_i(y) \Big \rangle\\
&=& \frac{1}{n^2} \sum_{\mathcal C_k} \frac{1}{q_{\mathcal C_k}} \sum_{i \neq j \in \mathcal C_k} \Big \langle \nabla f_i(x)-\nabla f_i(y),\nabla f_j(x)-\nabla f_j(y) \Big \rangle   \\
&\quad& + \frac{1}{n^2}\sum_{\mathcal C_k} \frac{1}{q_{\mathcal C_k}} \sum_{i \in \mathcal C_k}\frac{1}{p_i} \norm{\nabla f_i(x)-\nabla f_i(y)}^2\\
&=& \frac{1}{n^2} \sum_{\mathcal C_k} \frac{1}{q_{\mathcal C_k}} \norm{\sum_{i \in \mathcal C_k} \nabla f_i(x)-\nabla f_i(y)}^2  \\
&\quad& + \frac{1}{n^2}\sum_{\mathcal C_k} \frac{1}{q_{\mathcal C_k}} \sum_{i \in \mathcal C_k}\frac{1-p_i}{p_i} \norm{\nabla f_i(x)-\nabla f_i(y)}^2 \\
&\leq& \frac{1}{n^2} \sum_{\mathcal C_k} \frac{n_{\mathcal C_k}^2}{q_{\mathcal C_k}} 2L_{\mathcal C_k}D_{f_{\mathcal C_k}}(x, y)  \\
&\quad& + \frac{1}{n^2}\sum_{\mathcal C_k} \frac{1}{q_{\mathcal C_k}} \sum_{i \in \mathcal C_k}\frac{1-p_i}{p_i} 2L_iD_{f_i}(x,y) \\
&\leq& \frac{1}{n^2} \sum_{\mathcal C_k} \frac{n_{\mathcal C_k}^2}{q_{\mathcal C_k}} 2L_{\mathcal C_k}D_{f_{\mathcal C_k}}(x, y)  \\
&\quad& + \frac{1}{n^2}\sum_{\mathcal C_k} \frac{n_{\mathcal C_k}}{q_{\mathcal C_k}} 2\max_{i \in \mathcal C_k}\frac{1-p_i}{p_i}{L_i}D_{f_{\mathcal C_k}}(x,y) \\
&=& \frac{1}{n^2} \sum_{\mathcal C_k} 2(\frac{n_{\mathcal C_k}L_{\mathcal C_k}}{q_{\mathcal C_k}} + \max_{i \in \mathcal C_k}{\frac{(1-p_i)L_i}{q_{\mathcal C_k}p_i} }) n_{\mathcal C_k}D_{f_{\mathcal C_k}}(x, y)  \\
&\leq& 2\frac{1}{n} \max_{i \in \mathcal C_k}{(\frac{n_{\mathcal C_k}L_{\mathcal C_k}}{q_{\mathcal C_k}} + \max_{i \in \mathcal C_k}{\frac{(1-p_i)L_i}{q_{\mathcal C_k}p_i} )}} D_{f}(x,y).
\end{eqnarray*}

This gives the desired upper bound for the expected smoothness
\begin{eqnarray*}
\mathcal L(\tau) = \frac{1}{n}\max_{i \in \mathcal C_k}{\left(\frac{n_{\mathcal C_k}L_{\mathcal C_k}}{q_{\mathcal C_k}} + \max_{i \in \mathcal C_k}\frac{(1-p_i)L_i}{q_{\mathcal C_k}p_i} \right)}
.\end{eqnarray*}

Following the same notation, we move on to compute $\sigma$ for each sampling. First, for $\tau-$nice partition sampling we have 
\begin{eqnarray*}
\Exp{\norm{\nabla f_v(x^*)}^2} &=& \frac{1}{n^2}\sum_{\mathcal C_k} \sum_{i,j \in \mathcal C_k} \frac{\mathbf P_{ij}}{p_ip_j} \Big \langle \nabla f_i(x^*), \nabla f_j(x^*) \Big \rangle \\
&=& \frac{1}{n^2} \sum_{\mathcal C_k} \sum_{i \neq j \in \mathcal C_k} \frac{\mathbf P_{ij}}{p_ip_j} \Big \langle \nabla f_i(x^*),\nabla f_j(x^*) \Big \rangle   + \frac{1}{n^2} \sum_{\mathcal C_k} \sum_{i \in \mathcal C_k} \frac{1}{p_i} \Big \langle \nabla f_i(x^*),\nabla f_i(x^*) \Big \rangle\\
&=& \frac{1}{n^2} \sum_{\mathcal C_k} \sum_{i \neq j \in \mathcal C_k} \frac{n_{\mathcal C_k}(\tau-1)}{\tau(n_{\mathcal C_k}-1)q_{\mathcal C_k}} \Big \langle \nabla f_i(x^*),\nabla f_j(x^*) \Big \rangle   + \frac{1}{n^2} \sum_{\mathcal C_k} \sum_{i \in \mathcal C_k} \frac{n_{\mathcal C_k}}{\tau q_{\mathcal C_k}} \Big \langle \nabla f_i(x^*),\nabla f_i(x^*) \Big \rangle\\
&=& \frac{1}{n^2} \sum_{\mathcal C_k} \frac{n_{\mathcal C_k}(\tau-1)}{\tau(n_{\mathcal C_k}-1)q_{\mathcal C_k}} \sum_{i \neq j \in \mathcal C_k} \Big \langle \nabla f_i(x^*),\nabla f_j(x^*) \Big \rangle   + \frac{1}{n^2}\sum_{\mathcal C_k} \frac{n_{\mathcal C_k}}{\tau q_{\mathcal C_k}} \sum_{i \in \mathcal C_k} h_i\\
&=& \frac{1}{n^2} \sum_{\mathcal C_k} \frac{n_{\mathcal C_k}(\tau-1)}{\tau(n_{\mathcal C_k}-1)q_{\mathcal C_k}} \norm{\sum_{i \in \mathcal C_k} \nabla f_i(x^*)}^2  + \frac{1}{n^2}\sum_{\mathcal C_k} \frac{n_{\mathcal C_k}(n_{\mathcal C_k}-\tau)}{\tau(n_{\mathcal C_k}-1)q_{\mathcal C_k}}\sum_{i \in \mathcal C_k} h_i \\
&=& \frac{1}{n^2} \sum_{\mathcal C_k} \frac{n_{\mathcal C_k}^3(\tau-1)}{\tau(n_{\mathcal C_k}-1)q_{\mathcal C_k}} {h_{\mathcal C_k}}  + \frac{1}{n^2}\sum_{\mathcal C_k} \frac{n_{\mathcal C_k}^2(n_{\mathcal C_k}-\tau)}{\tau(n_{\mathcal C_k}-1)q_{\mathcal C_k}}\overline{h}_{\mathcal C_k}.
\end{eqnarray*}
Where its left to rearrange the terms to get the first result of the lemma. Next, we compute $\sigma$ for $\tau-$independent partition:

\begin{eqnarray*}
\Exp{\norm{\nabla f_v(x^*)}^2} &=& \frac{1}{n^2}\sum_{\mathcal C_k} \sum_{i,j \in \mathcal C_k} \frac{\mathbf P_{ij}}{p_ip_j} \Big \langle \nabla f_i(x^*), \nabla f_j(x^*) \Big \rangle \\
&=& \frac{1}{n^2} \sum_{\mathcal C_k} \sum_{i \neq j \in \mathcal C_k} \frac{\mathbf P_{ij}}{p_ip_j} \Big \langle \nabla f_i(x^*),\nabla f_j(x^*) \Big \rangle  + \frac{1}{n^2} \sum_{\mathcal C_k} \sum_{i \in \mathcal C_k} \frac{1}{p_i} \Big \langle \nabla f_i(x^*),\nabla f_i(x^*) \Big \rangle\\
&=& \frac{1}{n^2} \sum_{\mathcal C_k} \sum_{i \neq j \in \mathcal C_k} \frac{1}{q_{\mathcal C_k}} \Big \langle \nabla f_i(x^*),\nabla f_j(x^*) \Big \rangle  + \frac{1}{n^2} \sum_{\mathcal C_k} \sum_{i \in \mathcal C_k} \frac{1}{q_{\mathcal C_k}p_i} \Big \langle \nabla f_i(x^*),\nabla f_i(x^*) \Big \rangle\\
&=& \frac{1}{n^2} \sum_{\mathcal C_k} \frac{1}{ q_{\mathcal C_k}} \sum_{i \neq j \in \mathcal C_k} \Big \langle \nabla f_i(x^*),\nabla f_j(x^*) \Big \rangle   + \frac{1}{n^2}\sum_{\mathcal C_k} \frac{1}{q_{\mathcal C_k}} \sum_{i \in \mathcal C_k} \frac{1}{p_i}h_i\\
&=& \frac{1}{n^2} \sum_{\mathcal C_k} \frac{1}{q_{\mathcal C_k}} \norm{\sum_{i \in \mathcal C_k} \nabla f_i(x^*)}^2  + \frac{1}{n^2}\sum_{\mathcal C_k} \sum_{i \in \mathcal C_k} \frac{(1-p_i)h_i}{q_{\mathcal C_k}p_i} \\
&=& \frac{1}{n^2} \sum_{\mathcal C_k} \frac{n_{\mathcal C_k}^2}{q_{\mathcal C_k}} {h_{\mathcal C_k}} + \frac{1}{n^2}\sum_{\mathcal C_k} \sum_{i \in \mathcal C_k} \frac{(1-p_i)h_i}{q_{\mathcal C_k}p_i}.
\end{eqnarray*}

\section{Proof of Theorem \ref{le:tau_star}}
Recall that the optimal  batch size $\tau(x^*)$ is chosen such that the quantity $\max\left\{ \tau\mathcal L(\tau), \tfrac{2}{\epsilon\mu}\tau \sigma(x^*, \tau) \right\}$ is minimized. Note that in both $\tau-$nice partition, and $\tau-$ independent partition with $(p_i = \tfrac{\tau}{n_{\mathcal C_j}})$, $\tau \sigma(x^*, \tau)$ is a linear function of $\tau$ while $\tau \mathcal L(\tau)$ is a max across linearly increasing functions of $\tau$. To find the minimized in such a case, we leverage the following lemma.
\begin{lemma}
Suppose that $l_1(x),l_2(x),...,l_k(x)$ are increasing linear functions of $x$, and $r(x)$ is linear decreasing function of $x$, then the minimizer of $\max(l_1(x),l_2(x),...,l_k(x),r(x))$ is $x^* = \min_i(x_i)$ where $x_i$ is the unique solution for $l_i(x)=r(x)$
\end{lemma}
\emph{Proof:} Let $x^*$ be defined as above, and let $x$ be an arbitrary number. If $x \le x^*$, then  $r(x) \ge r(x^*) \ge r(x_i) = l_i(x_i) \ge l_i(x^*)$ for each $i$ which means $r(x) \ge \max(l_1(x^*), ..., l_k(x^*), r(x^*))$. On the other hand, if $x \ge x^*$, then let $i$ be the index s.t. $x_i=x^*$. We have $l_i(x) \ge l_i(x^*) = r(x^*) \ge r(x_j) = l_j(x_j) \ge l_j(x^*)$, hence $l_i(x) \ge \max(l_1(x^*), ..., l_k(x^*), r(x^*))$. This means that $x^* = \min_i(x_i)$ is indeed the minimizer of $\max(l_1(x),l_2(x),...,l_k(x),r(x))$.

Now we can estimate optimal  batch sizes for proposed samplings.

\textbf{$\tau-$nice partition:} if $\sum_{\mathcal C_j} \tfrac{n_{\mathcal C_j}^2}{e_j} ( h_{\mathcal C_j}^*n_{\mathcal C_j}-\overline {h}^*_{\mathcal C_j}) \le 0$ then $\tau \sigma(\tau)$ is a decreasing linear function of $\tau$, and $\tau \mathcal L(\tau)$ is the max of increasing linear functions. Therefore, we can leverage the previous lemma with $r(\tau) = \frac{2}{\epsilon \mu}\tau\sigma(x^*, \tau)$ and $l_{\mathcal C_k}(\tau) = \frac{n_{\mathcal C_k}^2(\tau-1)L_{\mathcal C_k} + n_{\mathcal C_k}(n_{\mathcal C_k}-\tau)\max_{i \in \mathcal C_k}{L_i} }{q_{\mathcal C_k}(n_{\mathcal C_k}-1)} n_{\mathcal C_k}$ to find the optimal  batch size as $\tau^* = \min\limits_{\mathcal C_k}(\tau^*_{\mathcal C_k})$, where 
\begin{equation*} 
    \tau^*_{\mathcal C_k} = \tfrac{
    \tfrac{nn_{\mathcal C_r}^2}{e_{\mathcal{C}_r}} (L_{\mathcal C_r}-L_{\max}^{\mathcal C_r}) + \tfrac{2}{\epsilon \mu} \sum_{\mathcal C_j} \tfrac{n_{\mathcal C_j}^3}{e_{\mathcal C_j}}\left(\overline {h}_{\mathcal C_j}- h_{\mathcal C_j}\right)}
    {\tfrac {nn_{\mathcal{C}_r}} {e_{\mathcal C_r}}(n_{\mathcal C_r}L_{\mathcal C_r}-L_{\max}^{\mathcal{C}_r}) + \tfrac{2}{\epsilon \mu}\sum_{\mathcal C_j} 
    \tfrac{n_{\mathcal C_j}^2 }{e_{\mathcal C_j}}({\overline {h}_{\mathcal C_j} -n_{\mathcal C_j}  h_{\mathcal C_j} })}.
\end{equation*}
\textbf{$\tau-$independent partition:} Similar to $\tau-$nice partition, we have $\tau \sigma(\tau)$ is a decreasing linear function of $\tau$ if $\quad\sum_{\mathcal C_j} \tfrac{n_{\mathcal C_j}}
{ q_{\mathcal C_j}} (n_{\mathcal C_j}  h_{\mathcal C_j}^* - \overline {h}^*_{\mathcal C_j}) \le 0$, and $\tau \mathcal L(\tau)$ is the max of increasing linear functions of $\tau$. Hence, we can leverage the previous lemma with $r(\tau) = \frac{2}{\epsilon \mu}\tau\sigma(x^*, \tau)$ and $l_{\mathcal C_k}(\tau) = {\frac{n_{\mathcal C_k}L_{\mathcal C_k}\tau}{q_{\mathcal C_k}} + (n_{\mathcal C_k}-\tau)\max_{i \in \mathcal C_k}{\frac{L_i}{q_{\mathcal C_k}} }}$ to find the optimal  batch size as $\tau^* = \min\limits_{\mathcal C_k}(\tau^*_{\mathcal C_k})$, where
\begin{equation*} 
    \tau^*_{\mathcal C_k} = {\tfrac{
    \tfrac{2}{\epsilon \mu}
    \sum_{\mathcal C_j} \tfrac{n_{\mathcal C_j}^2} { q_{\mathcal C_j}} \overline {h}_{\mathcal C_j} - \tfrac{n}{q_{\mathcal C_r}} L_{\max}^{\mathcal{C}_r} }
    {
    \tfrac{2}{\epsilon \mu}\sum_{\mathcal C_j} \tfrac {n_{\mathcal C_j}} { q_{\mathcal C_j}}
    (\overline {h}_{\mathcal C_j} - n_{\mathcal C_j} h_{\mathcal C_j})
    +
    \tfrac n {q_{\mathcal C_r}} (n_{\mathcal C_r}L_{\mathcal C_r}-L_{\max}^{\mathcal C_r}) 
    }}.
\end{equation*}

\section{Proof of bounds on step sizes}

For our choice of the learning rate we have
 
\begin{equation*}
\gamma^k = \frac 1 2\min \left\{\frac 1 {\lk}, \frac{\epsilon \mu}{\min(C,2\sg^k)} \right\}= \frac 1 2 \min \left\{\frac 1 {\lk}, \max \left\{ \frac{\epsilon \mu} C, \frac{\epsilon \mu}{2\sg^k} \right\} \right\} \geq  \frac 1 2\min \left\{\frac 1 {\lk}, \frac{\epsilon \mu} C  \right\}
.\end{equation*}
Since $\lk$ is a linear combination between the smoothness constants of the functions $f_i$, then it is bounded. Therefore,
both $\lk$ and $C$ are upper bounded and lower bounded as well as $\frac 1 {\lk}$ and $\frac{\epsilon \mu} C$, thus also $\gamma^k$ is bounded by positive constants $\gamma_{\max}= \tfrac 1 2\max_{\tau \in [n]} \left\{ \tfrac 1 {\mathcal L(\tau)} \right\}$ and $ \gamma_{\min}= \frac 1 2\min \left\{ \min_{\tau \in [n]} \left\{\tfrac 1 {\mathcal L(\tau)} \right\}, \tfrac{\epsilon \mu} C \right\} $.

\section{Proof of Theorem \ref{th:convergence_our}}
Let $r^k = \norm{x^k-x^*}^2$, then
\begin{eqnarray*}
\Exp{r^{k+1} |x^k} &=& \Exp{\norm{x^k-\gamma^k \nabla f_{v_k}(x^k)-x^*}^2 |x^k}\\
&=& r^k + (\gamma^k)^2\Exp{\norm{\nabla f_{v_k} (x^k)}^2|x^k} - 2 \gamma^k \langle \Exp{\nabla f_{v_k}(x^k)|x^k}, r^k \rangle  \\
&=&  r^k + (\gamma^k)^2 \Exp{\norm{\nabla f_{v_k}(x^k)}^2|x^k}-2\gamma^k \left(f(x^k)-f(x^*)+\frac \mu 2 r^k\right) \\
&=& (1-\gamma^k \mu)r^k+ (\gamma^k)^2 \Exp{\norm{\nabla f_{v_k} (x^k)}^2|x^k} - 2\gamma^k(f(x^k)-f(x^*))  \\
&\leq&  (1-\gamma^k \mu)r^k+ (\gamma^k)^2 (4 \mathcal L^k(f(x^k)-f(x^*)) +2\sg) - 2\gamma^k(f(x^k)-f(x^*))\\
&=& (1-\gamma^k \mu)r^k -2 \gamma^k((1-2\gamma^k \mathcal L^k)(f(x^k)-f(x^*))) +2(\gamma^k)^2\sg \\
&\leq& (1-\gamma^k \mu)r^k +2(\gamma^k)^2\sg \quad \text{ for  } \gamma_k \leq \frac 1 {2 \mathcal L^k} .\end{eqnarray*}

From Eariler bounds, there exist upper and lower bounds for step-sizes, $\gamma_{\min} \leq \gamma^k \leq \gamma_{\max}$, thus
\begin{equation*}
\Exp{r^{k+1} |x^k} \leq (1-\gamma^k \mu)r^k +2(\gamma^k)^2\sg  \leq (1-\gamma_{\min} \mu)r^k +2\gamma_{\max}^2\sg
.\end{equation*}
Therefore, unrolling the above recursion gives
\begin{eqnarray*}
\Exp{r^{k+1}|x^k} &\leq& (1-\gamma_{\min} \mu)^k r^0+ 2 \gamma_{\max}^2 \sg \sum_{i=0}^k(1-\gamma_{\min}\mu)^k \\
&\leq& (1-\gamma_\text{min} \mu)^{k}r^0+  \frac {2\gamma_{\max}^2\sg}{\gamma_\text{min} \mu}
.\end{eqnarray*}

\section{Proof of convergence to linear neighborhood in $\epsilon$}

In this section, we prove that our algorithm converges to a neighborhood around the optimal solution with size upper bounded by an expression linear in $\epsilon$.
First of all, we prove that $\sg(x)$ is lower bounded by a multiple of the variance in the optimum $\sg^*$ (in Lemma \ref{le:sg_lower_bound}). Then, by showing an alternative upper bound on the step-size, we obtain an upper bound for neighborhood size $R$ as expression of $\epsilon$.

\begin{lemma} \label{le:sg_lower_bound}
Suppose $f$ is  $\mu-$strongly convex, $L-$smooth and with expected smoothness constant $\mathcal L$. Let $v$ be as in the SGD overview, i.e., $\Exp{v_i=1}$ for all $i$.  Fix any $c>0$. The function $\sigma(x)= \Exp{\norm{\nabla{f_v(x)}}^2}$ can be lower bounded as follows: \[\sigma(x) \ge  \left\{ \mu^2 c   , 1 - 2 \sqrt{\mathcal L L c}  \right\}\sigma(x^*), \qquad \forall x\in \R.\]
The constant $c$ maximizing this bound is $c=\left(\frac{\sqrt{\mathcal L L +\mu^2} - \sqrt{\mathcal L L} }{\mu^2}\right)^2$, giving the bound
\[\sigma(x) \ge \frac{\left(\sqrt{\mathcal L L +\mu^2} - \sqrt{\mathcal L L} \right)^2}{\mu^2}  \sigma(x^*), \qquad \forall x\in \R.\]
\end{lemma}

{\em Proof:}
Choose $c>0$. If $\norm{x-x^*} \ge \sqrt{c\sigma(x^*)}$, then using Jensen's inequality and strong convexity of $f$, we get
\begin{equation}\label{eq:nif8g98gdud}
\sigma(x) \ge  \norm{\Exp{\nabla{f_v(x)}}}^2 
= \norm{\nabla{f(x)}}^2 
 = \norm{\nabla{f(x)}-\nabla{f(x^*)}}^2
\ge 
\mu^2 \norm{x-x^*}^2
 \ge 
\mu^2 c \sigma(x^*). 
\end{equation}

If $\norm{x-x^*} \le \sqrt{c\sigma(x^*)}$, then using expected smoothness and $L$-smoothness, we get
\begin{equation}
\Exp{\norm{\nabla{f_v(x)}-\nabla{f_v(x^*)}}^2}
\overset{\eqref{eq:exp_smoothn_cL}}{\leq} 2 \mathcal L (f(x)-f(x^*)) \leq  \mathcal L L  \norm{x-x^*}^2 \leq  \mathcal L L c\sigma(x^*). \label{eqL:bui7gd97dd}\end{equation}

Further, we can write
\begin{eqnarray*}
\sigma(x^*)-\sigma(x) &=& \Exp{\norm{\nabla{f_v(x^*)}}^2}-\Exp{\norm{\nabla{f_v(x)}}^2}\\
&=&
- 2\Exp{ \left\langle\nabla{f_v(x)}-\nabla{f_v(x^*)},\nabla{f_v(x^*)} \right\rangle } - \Exp{\norm{\nabla{f_v(x)}-\nabla{f_v(x^*)}}^2} \\
&\leq & - 2\Exp{ \left\langle\nabla{f_v(x)}-\nabla{f_v(x^*)},\nabla{f_v(x^*)} \right\rangle } \\
&\leq & 2\Exp{ \norm{\nabla{f_v(x)}-\nabla{f_v(x^*)}} \norm{ \nabla{f_v(x^*)}  }}\\
&\leq & 2 \sqrt{\Exp{ \norm{\nabla{f_v(x)}-\nabla{f_v(x^*)}}^2}} \sqrt{\Exp{\norm{ \nabla{f_v(x^*)}  }^2}}\\
&\overset{\eqref{eqL:bui7gd97dd}}{\le}&  2 \sqrt{\mathcal L L c} \sqrt{\sigma(x^*)}  \sqrt{\sigma(x^*)} \\
&=& 2 \sqrt{\mathcal L Lc} \sigma(x^*),\end{eqnarray*}
where the first inequality follows by neglecting a negative term, the second by Cauchy-Schwarz and the third by H\"{o}lder inequality for bounding the expectation of the product of two random variables.
The last inequality implies that 
\begin{equation}\label{eq:i9hfgdhuf}
\sigma(x) \ge \left(1 - 2 \sqrt{\mathcal L L c}  \right) \sigma(x^*) 
.\end{equation}
By combining  the bounds \eqref{eq:nif8g98gdud} and \eqref{eq:i9hfgdhuf}, we get
\[\sigma(x) \ge \min \left\{ \mu^2 c   , 1 - 2 \sqrt{\mathcal L L c}  \right\}\sigma(x^*).\]

\bigskip
\bigskip

Using Lemma \ref{le:sg_lower_bound}, we can upper bound step-sizes $\gamma^k$. Assume that $\sigma^*=\sigma(x^*)>0$. Let $\gamma_{\max}^{'}= \frac {\epsilon \mu} 2 \max \left\{ \frac 1 C, \frac 1 {2 \eta \sg^*} \right\}$, where $\eta = \frac{\left(\sqrt{\mathcal L L +\mu^2} - \sqrt{\mathcal L L} \right)^2}{\mu^2} $.

 We have
\begin{equation*}
\gamma^k = \frac 1 2\min \left\{\frac 1 {\lk}, \frac{\epsilon \mu}{\min(C,2\sg^k)} \right\} \leq \frac {\epsilon \mu} 2 \max \left\{ \frac{1} C, \frac{1}{2\sg^k} \right\} \leq \frac {\epsilon \mu} 2 \max \left\{ \frac{1} C, \frac{1}{2 \eta \sg^*} \right\} = \gamma_{\max}^{'}
.\end{equation*}

Now, we use this alternative step-sizes upper bound to obtain alternative expression for residual term $R= \frac {2\gamma^{2}_{\max}\sg^*}{\gamma_\text{min} \mu}$ in Theorem \ref{th:convergence_our} (let's denote it $R^{'}$). Analogically to proof of Theorem \ref{th:convergence_our} (with upper bound of step-sizes $\gamma_{\max}^{'}$), we have $R^{'}= \frac {2\gamma^{'2}_{\max}\sg^*}{\gamma_\text{min} \mu}$.

Finally, expanding expression for residual term $R^{'}$ yields the result: 
\begin{align*}
R^{'} &= \frac {2\gamma^{'2}_{\max}\sg^*}{\gamma_\text{min} \mu} = 
\frac {2\left(\frac {\epsilon \mu} 2 \max \left\{ \frac{1} C, \frac{1}{2 \eta \sg^*} \right\} \right)^2 \sg^*}{\frac 1 2\min \left\{ \min_{\tau \in [n]} \left\{\tfrac 1 {\mathcal L(\tau)} \right\}, \tfrac{\epsilon \mu} C \right\} \mu} =
\epsilon^2 \mu \left( \max \left\{ \frac{1} C, \frac{1}{2 \eta \sg^*} \right\} \right)^2 \max \left\{ \max_{\tau \in [n]} \left\{ \mathcal L(\tau) \right\}, \tfrac C {\mu \epsilon}\right\} \sg^*\\
&= \epsilon \mu \left( \max \left\{ \frac{1} C, \frac{1}{2 \eta \sg^*} \right\} \right)^2 \max \left\{ \epsilon \max_{\tau \in [n]} \left\{ \mathcal L(\tau) \right\}, \tfrac C \mu\right\} \sg^*
. \end{align*}
As conclusion, if we consider $R^{'}$ to be function of $\epsilon$, then it is at least linear in $\epsilon$,  $R^{'} \in \mathcal O (\epsilon)$.

\section{Additional Experimental Results}
\begin{figure}[ht]
\begin{center}
\centerline{
\includegraphics[width=0.25\columnwidth]{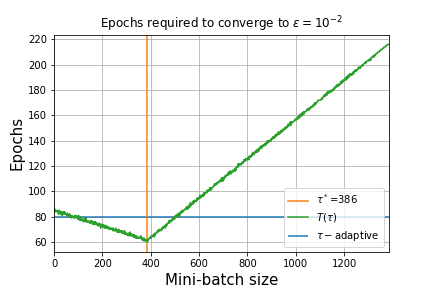}
\includegraphics[width=0.25\columnwidth]{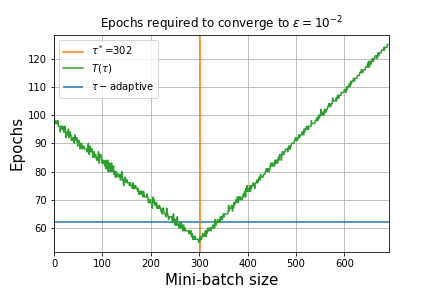}
\includegraphics[width=0.25\columnwidth]{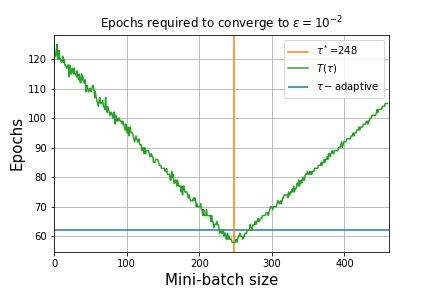}
\includegraphics[width=0.25\columnwidth]{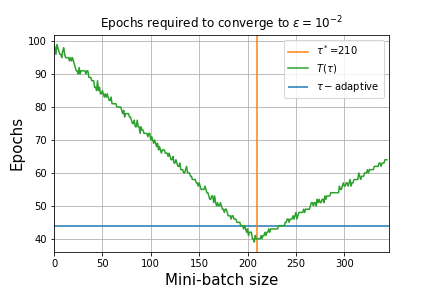}
}
\centerline{
\includegraphics[width=0.25\columnwidth]{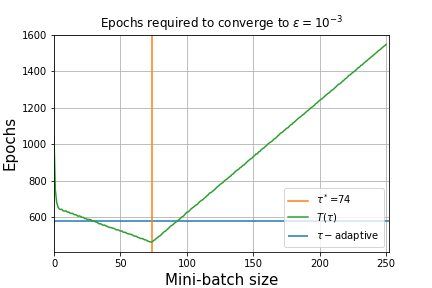}
\includegraphics[width=0.25\columnwidth]{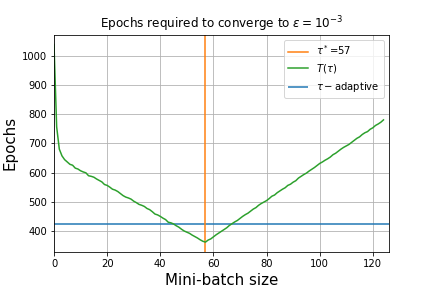}
\includegraphics[width=0.25\columnwidth]{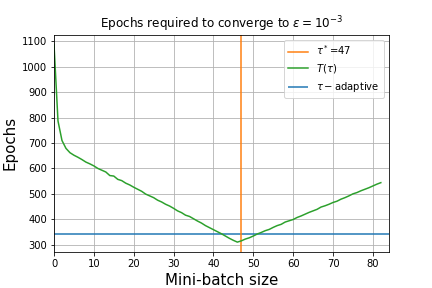}
\includegraphics[width=0.25\columnwidth]{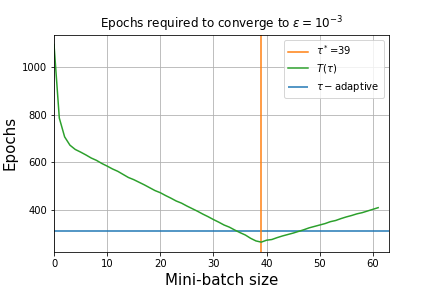}
}
\caption{\textbf{Effect of  batch size on the total iteration complexity}. First row: \emph{mg} dataset sampled using $\tau-$nice partition sampling. Second row: \emph{bodyfat} dataset sampled using $\tau-$independent partition sampling. From left to right: dataset is distributed across $1$, $2$, $3$, and $4$ partitions. This figure reflects the tightness of the theoretical result in relating the total iteration complexity with the  batch size, and the optimal  batch size. Moreover, This figure shows how close the proposed algorithm is to the optimal  batch size in terms of the performance.
}
\label{fig:taus_sup}
\end{center}
\vspace{-0.75cm}
\end{figure}

In this section, we present additional experimental results. Here we test each dataset on the sampling that was not tested on. Similar to the earlier result, the proposed algorithm outperforms most of the fixed  batch size SGD. Moreover, it can be seen as a first glance, that the optimal  batch is non-trivial, and it is changing as we partition the dataset through different number of partitions. For example, in \emph{bodyfat} dataset that is located in one partition, the optimal  batch size was $\tau^* = 74$. Although one can still sample $\tau = 74$ when the data is divided into two partitions, the optimal has changed to $\tau^* = 57$. This is clearly shown in Figure \ref{fig:taus_sup} where it shows that the optimal batch size varies between different partitioning, and the predicted optimal from our theoretical analysis matches the actual optimal.

\begin{figure}[ht]
\begin{center}
\centerline{
\includegraphics[width=0.25\columnwidth]{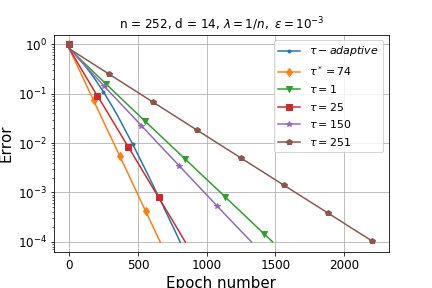}
\includegraphics[width=0.25\columnwidth]{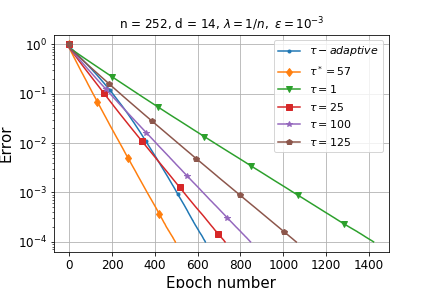}
\includegraphics[width=0.25\columnwidth]{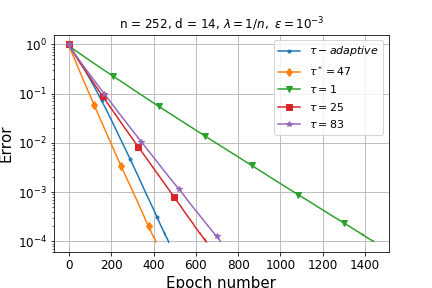}
\includegraphics[width=0.25\columnwidth]{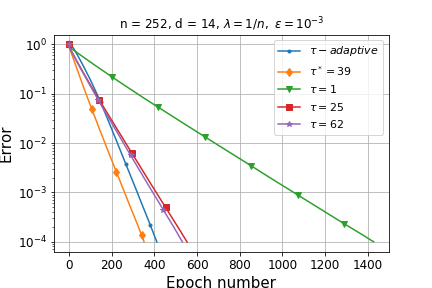}
}
\centerline{
\includegraphics[width=0.25\columnwidth]{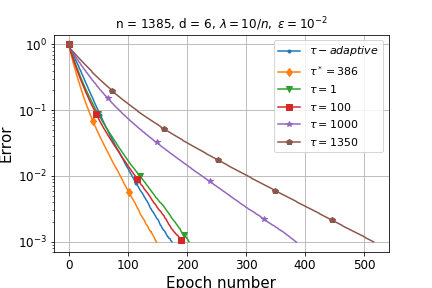}
\includegraphics[width=0.25\columnwidth]{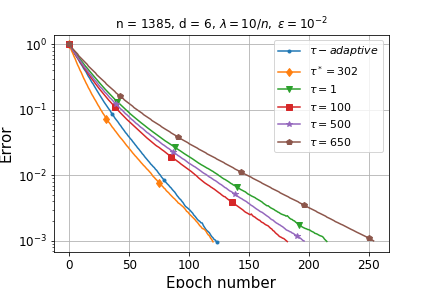}
\includegraphics[width=0.25\columnwidth]{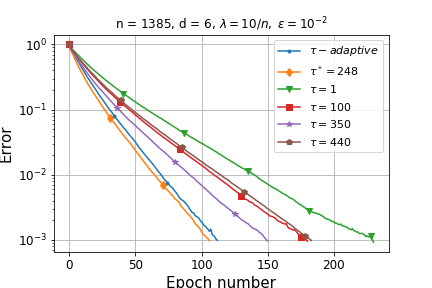}
\includegraphics[width=0.25\columnwidth]{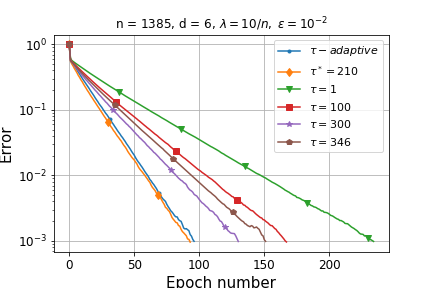}
}
\caption{\textbf{Convergence of Ridge regression} using $\tau-$partition independent sampling on \emph{Bodyfat} dataset (first row) and $\tau-$partition nice sampling on \emph{mg} dataset (second row). In first three columns, training set is distributed among $1$, $2$, $3$ and $4$ partitions, respectively.
}
\label{fig:ridge_sup}
\end{center}
\vspace{-0.75cm}
\end{figure}

\begin{figure}[ht]
\begin{center}
\centerline{
\includegraphics[width=0.25\columnwidth]{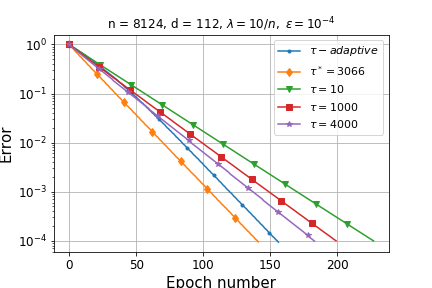}
\includegraphics[width=0.25\columnwidth]{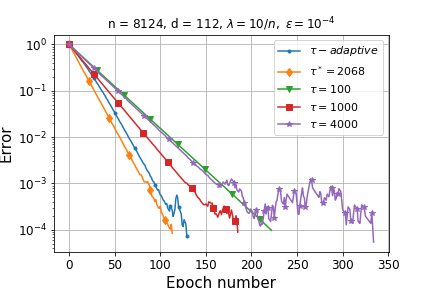}
\includegraphics[width=0.25\columnwidth]{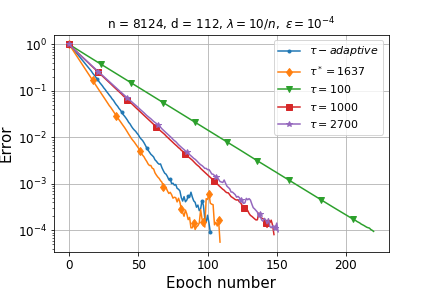}
\includegraphics[width=0.25\columnwidth]{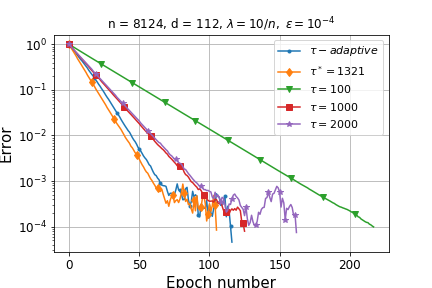}
}
\centerline{
\includegraphics[width=0.25\columnwidth]{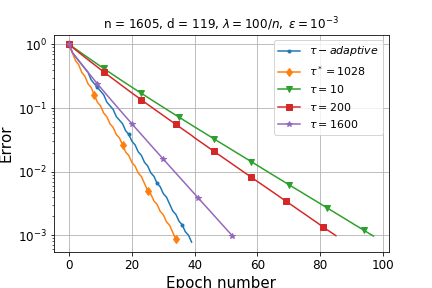}
\includegraphics[width=0.25\columnwidth]{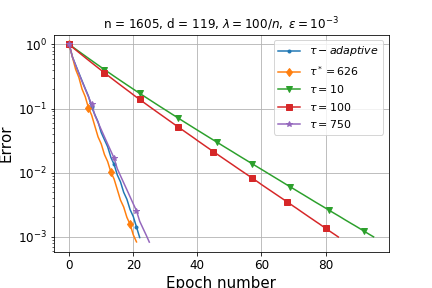}
\includegraphics[width=0.25\columnwidth]{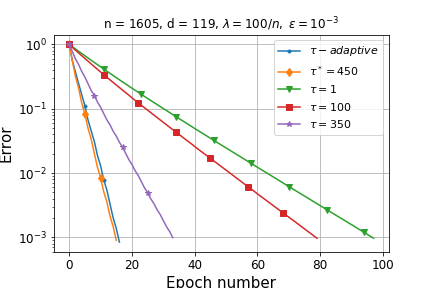}
\includegraphics[width=0.25\columnwidth]{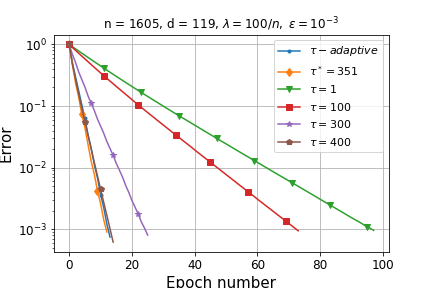}
}
\caption{\textbf{Convergence of Logistic regression} using $\tau-$partition independent sampling on \emph{mushrooms} dataset (first row) and $\tau-$partition nice sampling on \emph{a1a} dataset (second row). In first three columns, training set is distributed among $1$, $2$, $3$ and $4$ partitions, respectively.
}
\label{fig:logistic_sup}
\end{center}
\vspace{-0.75cm}
\end{figure}

\end{document}